\documentclass[letterpaper, 10pt, conference]{ieeeconf}
\IEEEoverridecommandlockouts

\usepackage{graphicx}
\usepackage{amsmath}
\usepackage{algpseudocode}
\usepackage{algorithm}
\usepackage{hyperref}
\usepackage{cleveref}
\usepackage{cite}
\usepackage{xcolor}
\usepackage{glossaries}
\usepackage{soul}

\newacronym{mpc}{MPC}{Model Predictive Control}
\newacronym{admm}{ADMM}{Alternating Direction Method of Multipliers}
\newacronym{rti}{RTI}{Real-Time Iteration}
\newacronym{ocp}{OCP}{Optimal Control Problem}
\newacronym{nlp}{NLP}{Nonlinear Programming}

\usepackage{xcolor}
\definecolor{lgray}{gray}{0.30}
\usepackage{eso-pic}

\usepackage{xcolor}
\usepackage{eso-pic}
\AddToShipoutPictureBG*{%
  \AtPageUpperLeft{%
    \setlength{\unitlength}{1mm}%
    \put(-9.5,-9){\makebox(\paperwidth,0)[c]{\parbox{0.9\textwidth}{2024 IEEE/RSJ International Conference on Intelligent Robots and Systems (IROS), pp. 12734-12741}}}
  }
}
\AddToShipoutPictureBG*{%
  \AtPageUpperLeft{%
    \setlength{\unitlength}{1mm}%
    \put(-9.5,-14){\makebox(\paperwidth,0)[c]{\parbox{0.9\textwidth}{DOI: \href{https://ieeexplore.ieee.org/document/10801676}{10.1109/IROS58592.2024.10801676}}}}
  }
}

\begin{document}

%\AddToShipoutPictureBG*{%
%  \AtPageUpperLeft{%
%    \makebox[\pdfpagewidth]{\raisebox{\dimexpr-\height-20pt}{%
%       \textcolor{lgray} {
%       Accepted for publication at the 2024 IEEE/RSJ International Conference on Intelligent Robots and Systems (IROS) }
%    }}%
%  }%
%}

\title{\LARGE \bf Accelerating Model Predictive Control for Legged Robots through Distributed Optimization}

\author{
Lorenzo Amatucci$^{1,2}$, 
Giulio Turrisi$^{1}$,
Angelo Bratta$^{1}$, 
Victor Barasuol$^{1}$, 
Claudio Semini$^{1}$ % <-this % stops a space
\thanks{$^{1}$ Dynamic Legged Systems Laboratory, Istituto Italiano di Tecnologia (IIT), Genova, Italy.
Email: {\tt\small name.surname@iit.it}}
\thanks{$^{2}$ Dipartimento di Informatica, Bioingegneria, Robotica e Ingegneria dei Sistemi (DIBRIS), Università di Genova, Genova, Italy.}
}

% make the title area
\maketitle

% As a general rule, do not put math, special symbols or citations
% in the abstract or keywords.

\begin{abstract}
This paper presents a novel approach to enhance Model Predictive Control (MPC) for legged robots through Distributed Optimization. Our method focuses on decomposing the robot dynamics into smaller, parallelizable subsystems, and utilizing the Alternating Direction Method of Multipliers (ADMM) to ensure consensus among them. Each subsystem is managed by its own Optimal Control Problem, with ADMM facilitating consistency between their optimizations. This approach not only decreases the computational time but also allows for effective scaling with more complex robot configurations, facilitating the integration of additional subsystems such as articulated arms on a quadruped robot. We demonstrate, through numerical evaluations, the convergence of our approach on two systems with increasing complexity. In addition, we showcase that our approach converges towards the same solution when compared to a state-of-the-art centralized whole-body MPC implementation. Moreover, we quantitatively compare the computational efficiency of our method to the centralized approach, revealing up to a 75\% reduction in computational time. Overall, our approach offers a promising avenue for accelerating MPC solutions for legged robots, paving the way for more effective utilization of the computational performance of modern hardware. Accompanying video at \url{https://youtu.be/Yar4W-Vlh2A}. The related code can be found at \url{https://github.com/iit-DLSLab/DWMPC}
\end{abstract}

%\IEEEpeerreviewmaketitle

\section{Introduction}

Planning and control for legged systems presents significant challenges due to their inherently nonlinear dynamics. Designing trajectories for such systems is a demanding task, as these trajectories must satisfy system constraints, including the robot's kinematic limits, actuator torque limits, and the constraints imposed by the surrounding environment, such as obstacles or gaps. \gls{mpc} \cite{MayneMPC} has emerged as a powerful technique for generating complex motions in robotic systems, particularly in the domain of legged robots \cite{Farbod2017}. \gls{mpc} approaches calculate the optimal control inputs based on the modeled system, resulting in complex constrained optimization problems that need to be solved rapidly enough to achieve online re-planning of the optimal trajectory.
Recent advancements in onboard hardware computational power and the development of more efficient \gls{nlp} solvers like \cite{crocoddyl},\cite{acados}, and \cite{ocs2} have opened doors to solving efficiently complex \gls{ocp}. However, the solution time is still a crucial factor to be considered when designing a control architecture for legged systems.
To mitigate the computational complexity associated with the whole-body dynamics of the robot and speed up the solving time of the \gls{ocp}, the works \cite{Seungwoo20}, \cite{romualdi22}, and \cite{bratta20} opted to approximate the robot with the centroidal model~\cite{Orin}. This approach reduces the robot's dynamics to its center of mass (CoM) motion, neglecting the influence of individual limbs on the base. Although this approximation may seem simplistic, it considers the key dynamic effects on the system. Additionally, it facilitates the integration of feasibility constraints, such as contact friction.
The centroidal model has demonstrated effectiveness in stabilizing locomotion for both quadrupeds and bipeds, although it is necessary to carefully craft reference trajectories to track agile motions.

\begin{figure}[!t]
%\vspace{5pt}
\centering{
\includegraphics[width=0.48\textwidth]{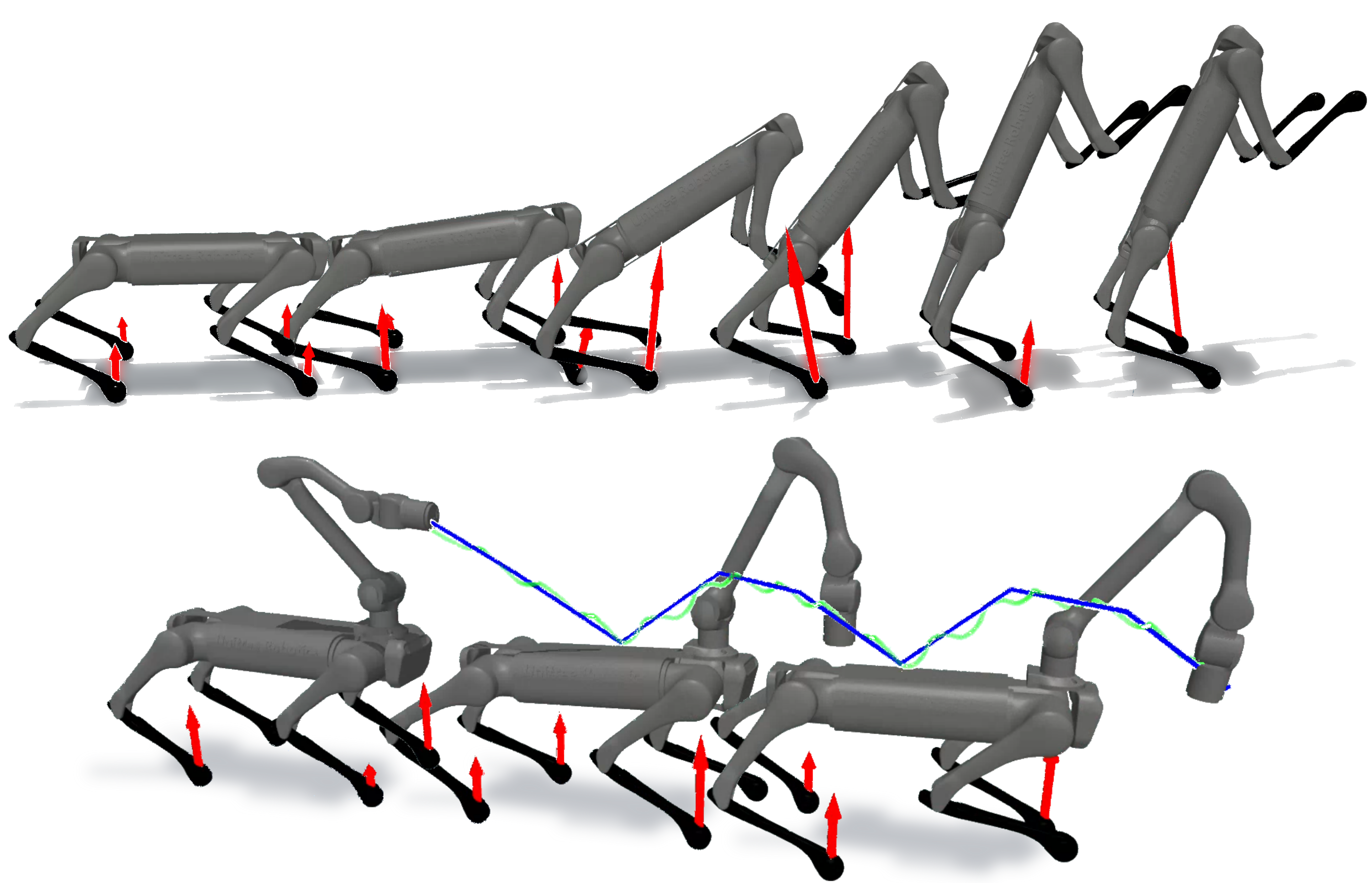}
}    
\caption{Simulation snapshots of robotic systems controlled by the proposed MPC with distributed optimization, performing different agile motions. On the top, a quadruped standing up on two feet and walking forward. On the bottom, a quadruped manipulator following a triangular spiral with the manipulator end-effector, the reference is highlighted in blue while the actual trajectory is in green.}
\label{fig:agility}
\end{figure}

To reduce the drawbacks of such a model, \cite{Whole_body_loco_manipulation_MPC} and \cite{Perceptive_based_MPC} developed an \gls{mpc} formulation based on the Kino-Dynamic model, which takes into account joint position and velocity, neglects the effect of limb acceleration on the base dynamics while maintaining the effect of the robot joint configuration on the inertia matrix. As a result, the Kino-Dynamic model avoids the highly non-linear terms while simultaneously providing a better approximation than the centroidal model.
In contrast, \cite{inverse_dynamics_mpc} and \cite{talosWholeBody} used the full-body model, entirely exploiting the robot's capabilities. 
The complexity of the \gls{ocp}s considered in those implementations is remarkable, given the limited time budget available for solving them in the feedback \gls{mpc} scheme. Indeed, it needs to be considered that the computational complexity of the \gls{nlp} implementation increases with the number of state variables in the \gls{ocp}, typically following a cubic $O(n^3)$ law \cite{accelerating_ddp}.
This poses a significant limitation on the number of variables that can be feasibly included in the optimization process and restricts the length of the \gls{mpc} horizon. Despite the continuous growth of the number of CPU cores of each generation of modern processors, the algorithms presented earlier face limitations in terms of parallelizability. This limits the effective utilization of the computational resources and restricts how the solver's solution time scales with the number of decision variables. A different paradigm that tries to reduce the computation burden is presented in \cite{coupled_control_system}. The authors worked towards a reduction of the computational effort by decoupling a quadruped robot into two bipeds. This brought a consistent reduction in the number of optimization variables that drastically decreased the computational time; nevertheless, the formulation they employed limits the approach to only the offline generation of periodic orbits, too slow to be computed online in a feedback \gls{mpc} scheme. In contrast, \cite{distributedQP} proposed a similar decoupling principle and applied it to quadratic programming-based nonlinear controllers. While the parallelizability of their approach leads to remarkable performances, particularly in terms of disturbance rejection properties inherited from the decoupling framework, their implementation requires the knowledge of the centralized system optimum. 
The approach is thus limited to tracking pre-computed periodic orbits. Finally, \cite{coupled_control_LF} developed local controllers based on the dynamic decomposition from \cite{coupled_control_system}, which can stabilize the centralized system across various scenarios. However, their approach is not suitable for generating highly dynamic motions due to the intrinsic limitations of instantaneous controllers. The above-mentioned methods also lack easy adaptability to more complex systems, such as a quadruped robot with one or several articulated arms on top.

Our approach divides the robot into multiple subsystems while ensuring coherence between the solutions of the decoupled systems with a consensus formulation. In particular, we designed an easily scalable method to speed up the computation of the \gls{ocp} on legged systems exploiting a parallelized implementation of the \gls{admm}. It is worth highlighting that \gls{admm} has already been successfully used in legged robotics by \cite{dynamics_consensus}. Their work primarily focused on rigorously enforcing consensus between simplified and whole-body optimizations within a two-step optimization framework, which is common in locomotion controllers.
In contrast, our approach focuses on decomposing the robot dynamics into reduced subsystems and using \gls{admm} to ensure consensus among them. In other words, our approach boils down to running in parallel a separate \gls{mpc} for each subsystem we have divided our robot into, using consensus \gls{admm} to maintain consistency between the optimizations. This allows our algorithm to scale effectively with more complex robot configurations. For instance, the issue of integrating an articulated arm onto a quadruped is simplified to merely adding another subsystem in parallel to the others and adding the relative consensus. By adopting this methodology, our method is able to solve complex whole-body motions (Fig. \ref{fig:agility}), overcoming at the same time the curse of dimensionality that is normally associated with the use of the robot's full dynamics.

%, and maintaining consistent solving times even for more challenging systems.

\begin{figure}[!t]
%\vspace{5pt}
\centering{
\includegraphics[width=0.3\textwidth]{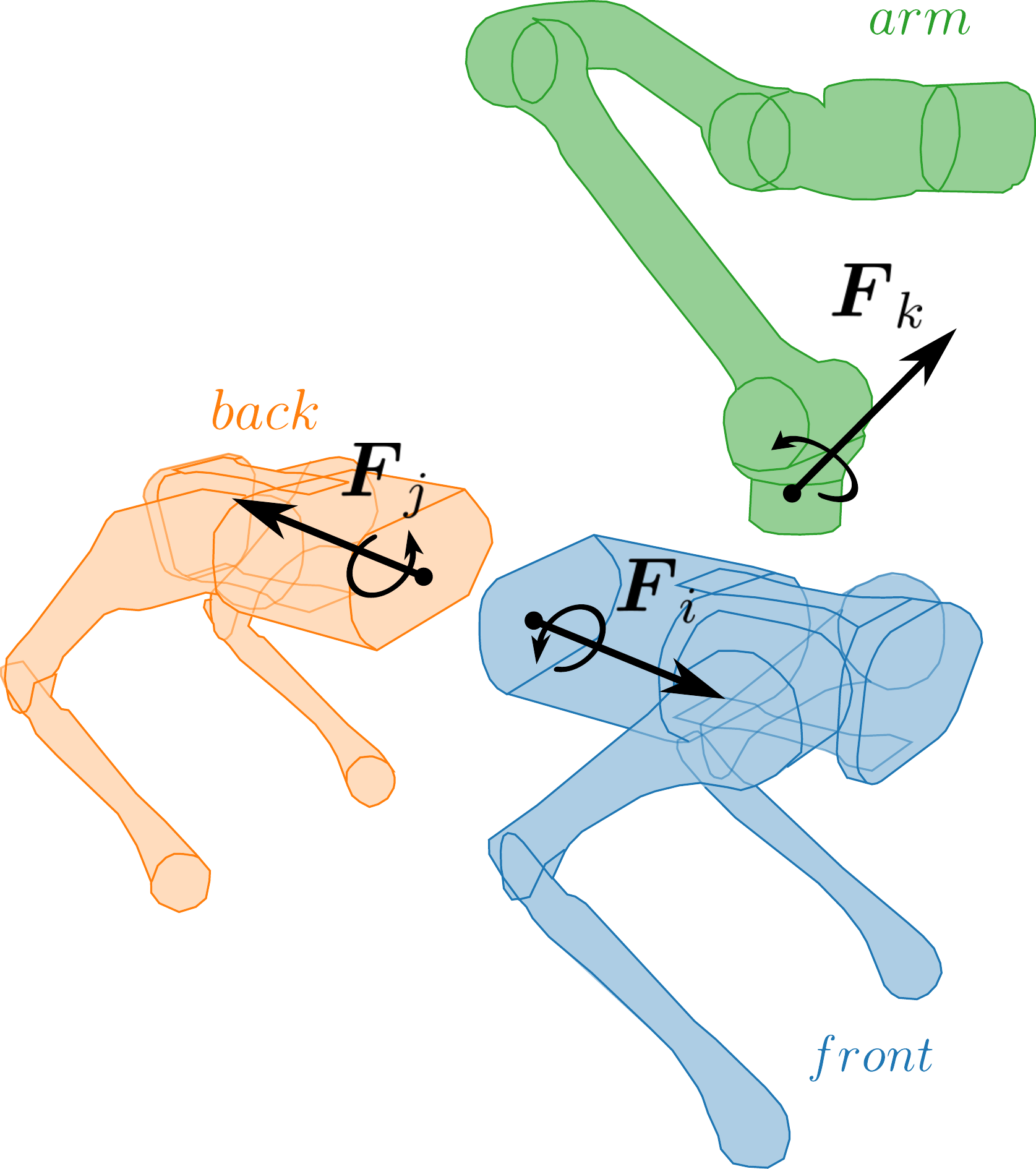}
}    
\caption{Schematic of a quadruped robot with an articulated arm on top split into three separate subsystems. Each section of the robot sees the dynamic effect of the other parts through the interaction wrench $\boldsymbol{F}$.}
\label{fig:robot_splitted}
\vspace{-18pt}%ADL
\end{figure}
\subsection{Contributions}
The main contributions of this work are:
\begin{itemize}
    \item the development of a novel scalable framework based on \gls{admm} that reduces the computational time of the \gls{ocp} problem, e.g. two times reduction for a quadruped robot and up to four times for a quadruped plus a 6 Degrees of Freedom (DoF) arm on top, Fig. \ref{fig:robot_splitted}.
    \item a numerical evaluation of the scalability of the proposed approach to systems of increasing complexity, e.g. adding an arm on a quadruped robot does not affect the computational time.
    \item a quantitative evaluation, in simulation of the performance and convergence property of the proposed method against a state-of-the-art implementation of the same whole-body \gls{mpc} problem.
\end{itemize}

\subsection{Outline}
This paper is organized as follows. Section \ref{sec::background} presents the necessary backgrounds on the \gls{admm} algorithm and \gls{mpc}. Section \ref{sec::DWMPC} describes our approach in detail. Section \ref{subsec::convergence} demonstrates the convergence property of our formulation, while Section \ref{subsec::performance} shows the performance in terms of computational time of our distributed whole body \gls{mpc}, in a simulation environment. Finally, Section \ref{sec::conclusion} draws the final considerations and conclusions.

\section{Background}
\label{sec::background}
\subsection{Alternating Direction Method of Multipliers}
\label{subsec::ADMM}
\gls{admm} is an efficient approach for solving optimization problems. It has been broadly adopted in both robotics and machine learning fields \cite{admm_Boyd1},\cite{admm_robotics}. The \gls{admm} algorithm solves problems in the form:
\begin{equation}  
\label{eq::admm_op}
\begin{aligned}
&  \min\limits_{\boldsymbol{w},\boldsymbol{z}} \quad \boldsymbol{f}(\boldsymbol{w}) + \boldsymbol{g}(\boldsymbol{z}) \\
& \text{s.t.} \quad \boldsymbol{A}\boldsymbol{w} + \boldsymbol{B}\boldsymbol{z} = \boldsymbol{c} 
\end{aligned}
\end{equation}
  
Where $\boldsymbol{A}$,$\boldsymbol{B}$, and $\boldsymbol{c}$ are respectively the matrices and the vector that define the equality constraint.
$\boldsymbol{w}$ and $\boldsymbol{z}$ are two separate decision variables. $\boldsymbol{f}(\boldsymbol{w})$ and $\boldsymbol{g}(\boldsymbol{z})$ are the two objectives of the cost function. The augmented Lagrangian for \eqref{eq::admm_op} is:
\begin{equation}  
\label{eq::lagrangian}
\begin{split}
    \mathcal{L}_{\rho}(\boldsymbol{w}, \boldsymbol{z}, \boldsymbol{y}) = &\boldsymbol{f}(\boldsymbol{w}) + \boldsymbol{g}(\boldsymbol{z}) + \boldsymbol{y}^T(\boldsymbol{A}\boldsymbol{w} + \boldsymbol{B}\boldsymbol{z} - \boldsymbol{c}) + \\ &\frac{\rho}{2}||\boldsymbol{A}\boldsymbol{w} + \boldsymbol{B}\boldsymbol{z} - \boldsymbol{c}||^2
\end{split}
\end{equation}
where $\boldsymbol{y}$ is the dual variable associated with the equality constraint and $\rho$ is the penalty parameter related to the constraint violation. 
The \gls{admm} exploits the separation of the cost terms between the $\boldsymbol{w}$ and $\boldsymbol{z}$ decision variables to split the problem and thus tackling smaller and simpler sub-problems at each iteration \cite{admm_Boyd1}.
The solution to \eqref{eq::admm_op} can be found iterating till convergence the following:
\begin{equation}
\label{eq::admm_iteration}
\begin{aligned}
\boldsymbol{w}^{n+1} & =  \min\limits_{\boldsymbol{w}} \mathcal{L}^{\rho}(\boldsymbol{w}, \boldsymbol{z}^n, \boldsymbol{y}^n) \\
\boldsymbol{z}^{n+1} & =  \min\limits_{\boldsymbol{z}} \mathcal{L}_{\rho}(\boldsymbol{w}^{n+1}, \boldsymbol{z}, \boldsymbol{y}^n) \\
\boldsymbol{y}^{n+1} & = \boldsymbol{y}^n + \rho(\boldsymbol{A}\boldsymbol{w} + \boldsymbol{B}\boldsymbol{z} - \boldsymbol{c})
\end{aligned}
\end{equation}
where the superfixed $n$ indicates the iteration number. 
In the first two steps of \eqref{eq::admm_iteration}, we update the two primal variables $\boldsymbol{w}$ and $\boldsymbol{z}$ sequentially, and then we calculate the dual variable;  
this results in a sequential implementation of the algorithm.\\
A widely adopted approach performs the updates of primal variables, $\boldsymbol{w^{n+1}}$ and $\boldsymbol{z}^{n+1}$ in parallel. Parallelization is achieved at the expense of both the convergence rate and convergence guarantees, prioritizing on the other hand computational speed. 
~\cite{parallel_admm_convergence} showed that including additional regularization terms in the cost partially recovers convergence rate and guarantees. 
In our method, we will use a special case of \gls{admm} called \textit{consensus} \gls{admm}~\cite{admm_Boyd1}, in which the problem is in the form:
\begin{equation}
\begin{aligned}
    &\min_{\boldsymbol{\Bar{w}}, \boldsymbol{w}_1, \boldsymbol{w}_2, \ldots, \boldsymbol{w}_{N_{sys}}} \sum_{i=1}^{N_{sys}} \boldsymbol{f}_i(\boldsymbol{w}_i) \\
&\text{s.t. } \boldsymbol{w}_i = \boldsymbol{\Bar{w_i}}, \quad i = 1, \ldots, N_{sys} \\
\end{aligned}
\end{equation}
where $N_{sys}$ is the number of subsystems we split the main problem into. $\boldsymbol{w}_i$ is a local copy of a subvector of the global decision variable $\boldsymbol{\Bar{w}}$. 
The objective in this formulation is to optimize $\boldsymbol{\Bar{w}}$ through the separate optimization of the $\boldsymbol{w}_i$ variables. In this way, we transform the centralized optimization involving only  $\boldsymbol{\Bar{w}}$ into a network of smaller local independent problems. In our implementation, we will divide the robot model into separate independent \gls{ocp}, which can be solved in parallel. The decision variables and constraints of the global problem are reduced to the ones related to the considered subsystems, making the problem much smaller. The consensus is then established between the optimizations, to avoid non-physical behaviors.\\
\subsection{Model Predictive Control}
\label{subsec::MPC}
In this section, we will first briefly introduce the dynamic model we are using and then we will present the \gls{ocp}. The efficacy of the control input derived from an \gls{mpc} strategy is directly linked with the accuracy of its predictions. Carefully modeling the dynamics of the robot is necessary to maximize its performance. Therefore, we have chosen to employ the articulated rigid body dynamics\cite{Roy} with hard contacts to characterize the dynamics of a legged robot accurately.
We define the state of the robot at instant $k$ by the generalized coordinates $\boldsymbol{q}_k \in {R}^{n_q}$ and the generalized velocity $\boldsymbol{v}_k \in {R}^{n_v}$.
We write the equation of motion of legged robots in discrete time following a semi-implicit Euler scheme as:
\begin{equation}
\label{eq::wholebody}
\begin{split}
&\boldsymbol{v}_{k+1} = \boldsymbol{v}_{k} - \boldsymbol{M}^{-1}(\boldsymbol{b} - \boldsymbol{S}\boldsymbol{\tau} - \boldsymbol{J}^T\boldsymbol{\lambda})dt \\
&\boldsymbol{q}_{k+1} = \boldsymbol{q}_{k} \oplus \boldsymbol{v}_{k+1}dt
\end{split}
\end{equation}
where $\boldsymbol{M} \in R^{n_v \times n_v}$ and $\boldsymbol{b} \in R^{n_v}$ 
are respectively the mass matrix and the "bias" vector that includes the Coriolis, centrifugal, and gravitational terms. The dependency of $\boldsymbol{M} $ and $ \boldsymbol{b}$ from $\boldsymbol{q}_k$ and $\boldsymbol{v}_k$ has been dropped in favor of readability. The variable $\boldsymbol{\tau} \in R^{n_u}$ corresponds to the joint torque, while $\boldsymbol{S} \in R^{n_v \times n_u}$ is a selector matrix. Matrix $\boldsymbol{J} \in R^{n_v \times n_c}$ is the stack of jacobian associated with each contact and $\boldsymbol{\lambda} \in R^{n_c}$ is the stack of ground reaction forces. Considering the dynamics described before, we can write an \gls{mpc} that solves the following \gls{ocp} in a receding horizon fashion:
\begin{subequations}
    \label{eq::ocp}
    \begin{align}
        \min\limits_{\boldsymbol{x},\boldsymbol{u}} \quad &\boldsymbol{l}_T(\boldsymbol{x}_N)+\sum_{k = 0}^{N-1}{\boldsymbol{l}(\boldsymbol{x}_k ,\boldsymbol{u}_k)}\\
         s.t. \quad &\boldsymbol{v}_{k+1} = \boldsymbol{v}_{k} - \boldsymbol{M}^{-1}(\boldsymbol{b} - \boldsymbol{S}\boldsymbol{\tau} - \boldsymbol{J}^T\boldsymbol{\lambda})dt \label{sub_eq::ocp_dynamcis} \\
&\boldsymbol{q}_{k+1} = \boldsymbol{q}_{k} \oplus \boldsymbol{v}_{k+1} dt \label{sub_eq::ocp_integrator}\\
        &\boldsymbol{J}\boldsymbol{v}_k=\boldsymbol{v_e}\label{sub_eq::ocp_holo}\\
        &\boldsymbol{p}_k|_z - \boldsymbol{p_{ref}}_k|_z = 0 \label{sub_eq::ocp_z_leg}\\
         &\boldsymbol{u}_k \; \in \; \text{U}_k \label{sub_eq::ocp_uk}\\
         &\boldsymbol{x}_k \; \in \; \text{X}_k\label{sub_eq::ocp_xk}\\
        &\boldsymbol{x}_{0} = \boldsymbol{\hat{x}}_
        {0}\label{sub_eq::ocp_x0}
    \end{align}
\end{subequations}
where $\boldsymbol{x}_k = \left[ \boldsymbol{v}_k , \boldsymbol{q}_k \right]^T \in R^{n_v+n_q}$ and $\boldsymbol{u}_k = \left[ \boldsymbol{\tau}_k , \boldsymbol{\lambda}_k \right]^T \in R^{n_u+n_c}$ are respectively the state and control input. $\boldsymbol{l}(\boldsymbol{x}(\cdot))$ is a quadratic cost including a tracking term and a regularization one.
$\boldsymbol{J}\boldsymbol{v}_k=\boldsymbol{v}_e$ slacks the non-slipping condition for each contact to avoid numerical instability \cite{BAUMGARTE}. $\boldsymbol{p}_k|_z$ is the position of the foot along the z-axis and it is constrained to guarantee that the legs in stance are touching the ground. $\text{U}_k$ is the set of feasible control input limits by the friction cone and torque limits at the joint. The set $\text{X}_k$ represents the feasible states constrained by the joint kinematic limits. $\boldsymbol{\hat{x}}_0$ is the state initial condition.\\
The described nonlinear optimization problem poses significant challenges when used for controlling legged robots since, due to their complexity, they are difficult to solve at high frequencies. Even with the latest advancements highlighted in \cite{inverse_dynamics_mpc}, the update rate of the final \gls{mpc} is limited to 50Hz. Our approach aims to enhance the capabilities of modern solvers by diving the \gls{ocp} into independent problems with a reduced number of decision variables that are solved in parallel. For instance, decoupling a quadruped into two parts, as illustrated in Fig. \ref{fig:robot_splitted}, results in a reduction with respect to the centralized approach by ~25\% of the number of variables in each subsystem's optimization. 
While the number of variables of the system incorporating also the additional arm on top is reduced by ~40\%. More details on the computational time gains derived from this reduction are provided in Section \ref{subsec::performance}. %We will now proceed to outline our algorithm formulation in detail.
\section{Distibuted Whole Body MPC}
\label{sec::DWMPC}
The main contribution of this work is to leverage the efficiency and parallelizability of the consensus \gls{admm} to speed up the solution time of the \gls{nlp} in \eqref{eq::ocp}. Our method provides a framework for partitioning the original \gls{ocp} into smaller, independent problems while ensuring consensus among the solutions. Through this approach, we can significantly decrease the computational time while augmenting our system's complexity, such as incorporating an articulated arm onto a quadruped robot, without significantly impacting the solution time, as reported in Section \ref{subsec::performance}. We can effectively decouple the robot into distinct subsystems, as illustrated in the example in Fig. \ref{fig:robot_splitted}. The consensus \gls{admm} implementation is used to enforce the consistency of the dynamic between the different subsystems and prevent them from drifting apart. Without consensus, we will only have decoupled independent systems that are no longer solving together the centralized \gls{ocp} in \eqref{eq::ocp}.

\subsection{Decomposed dynamics}
The decomposition of the robot dynamics is based on the idea of considering the separated subsystems interacting with each other through a wrench $\boldsymbol{F}$.
For the $i^{th}$ subsystem the wrench $\boldsymbol{F}_i$ synthesizes the dynamic effect of the other $j$ subsystems that we are splitting the robot into. The equations of motion of the $i^{th}$ subsystem also considering the rigid body connection constraint are written as:
\begin{subequations}
\label{eq::decomposed_dynamics}
\begin{align}
&\boldsymbol{v}_{i,k+1} = \boldsymbol{v}_{i,k} - \boldsymbol{M}^{-1}_i(\boldsymbol{b}_i - \boldsymbol{S}\boldsymbol{\tau}_i - \boldsymbol{J}_i^T\boldsymbol{\lambda}_i - \boldsymbol{J}_{F,i}^T\boldsymbol{F}_i)dt \label{sub_eq::dyn} \\
&\boldsymbol{J}_{F,i}\boldsymbol{v}_{i,k} = \boldsymbol{J}_{F,j}\boldsymbol{v}_{j,k} \quad \forall j \neq i\label{sub::eq_holo}
\end{align}
\end{subequations}
for $i,j \in \mathcal{X}$ where $\mathcal{X}$ is the set of subsystems we have divided our robot into.  $\boldsymbol{J}_{F,i}$ is the Jacobian that maps the generalized velocity of the subsystems into the velocity at the interface with the other systems. Manipulating \eqref{eq::decomposed_dynamics} we can write $\boldsymbol{F}_i$ as:
\begin{equation}
\label{eq::F}
      \boldsymbol{F}_i dt= (\boldsymbol{\Bar{J}}_{F}\boldsymbol{\Bar{M}}^{-1}\boldsymbol{\Bar{J}}_{F}^T)^{-1}\Delta v_{tot,k}
\end{equation}
Here, $\boldsymbol{\Bar{M}} = diag(M_0,\:\ldots \:,M_{N_{sys}})$ and $\boldsymbol{\Bar{J}}_F = \left[ \boldsymbol{J}_{F,0}\:\ldots \: \boldsymbol{J}_{F,N_{sys}} \right]^T$. $\Delta v_{tot}$ represent the sum of the effect of all systems and is defined as: 
\begin{equation}
\label{eq::deltav}
    \Delta v_{tot} \equiv \sum_{* \in \mathcal{X}} \boldsymbol{J}_{F,*}\left(\boldsymbol{v}_{*,k} + \boldsymbol{M}^{-1}_*(\boldsymbol{b}_* - \boldsymbol{S}\boldsymbol{\tau}_* - \boldsymbol{J}_*^T\boldsymbol{\lambda}_*)dt 
    \right)
\end{equation}
 It is worth noting that $\Delta v_{tot}$ is obtained by applying the dynamic of \eqref{eq::decomposed_dynamics} to each subsystem composing the considered body.
 The obtained coupling wrench in \eqref{eq::F} depends on all subsystem's generalized velocity and coordinates. Given the parallelization employed in this work, it is not possible to have the $j^{th}$ subsystems variables beforehand. This forced us to utilize the past information at algorithm iteration $n$ when calculating the $n+1$ update for the $i^{th}$ subsystem as done in the \gls{admm}. 
 Such an approximation, deriving from the subsystem decoupling, does not affect the optimization performances, as empirically demonstrated in section \ref{subsec::convergence}. Indeed, such an approximation is comparable to the one already used in centralized MPC employing Newton-base \gls{nlp} solvers, which linearized at each iteration based on the previous optimal solution \cite{Bemporad}.
Finally, the subsystem dynamics in \eqref{sub_eq::dyn} can be rewritten explicating the dependency of $\boldsymbol{F}_i$ as:
\begin{equation}
\label{eq::subsystem_dynamics}
\begin{aligned}
    \boldsymbol{v}^{n+1}_{i,k+1} = &\boldsymbol{v}^{n+1}_{i,k} - \boldsymbol{M}^{-1}_i(\boldsymbol{b}_i - \boldsymbol{S}\boldsymbol{\tau}^{n+1}_i - \boldsymbol{J}_i^T\boldsymbol{\lambda}^{n+1}_i + \\ 
    - \boldsymbol{J}_{F,i}^T\boldsymbol{F}&(\boldsymbol{q}_{i}^{n+1},\boldsymbol{v}_{i}^{n+1},\boldsymbol{\lambda}_{i}^{n+1},\boldsymbol{\tau}_{i}^{n+1},\boldsymbol{q}_{j}^n,\boldsymbol{v}_{j}^n,\boldsymbol{\lambda}_{j}^n,\boldsymbol{\tau}_{j}^n))dt   
\end{aligned}
\end{equation}
Equation \eqref{eq::subsystem_dynamics} now only depends on the $i^{th}$ system current decision variables, while the $j^{th}$ subsystem's influence only appears as a parameter coming from the last $n^{th}$ iteration. The system dynamics are reduced to only the floating base plus the joints considered in the $i^{th}$ subsystem. We have lost the direct dependency on the other joints, but we gained a model of reduced size that can be utilized in our distributed \gls{mpc}.
\subsection{Distributed optimization}
In this subsection, we will delve into the formulation of the distributed \gls{ocp}. The peculiarity of such implementation lies in the fact that we are going to solve \eqref{eq::ocp} defining $N_{sys}$ optimization problems, one per each subsystem we are dividing the robot into. Such problems are smaller than the problem of \eqref{eq::ocp}, and can be solved in parallel, thus speeding up the computation. Each subsystem optimization problem is defined utilizing the system dynamic in \eqref{eq::subsystem_dynamics} and the constraint \eqref{sub::eq_holo}. Such constraint serves to ensure coherence between velocities at the interface of the subsystems, thereby preventing them from drifting apart. Since we are decoupling the subsystems to enable parallel optimization, analogously to what we did for the decomposed dynamics, \eqref{sub::eq_holo} cannot be directly used, but it can be formulated as a consensus problem.
Defining the residual between the subsystems $i$ and $j$ at the time $k$ as :
\begin{equation}
\label{residual}
    \boldsymbol{r}_{i,j} = \boldsymbol{J}_{F,i} \boldsymbol{v}_i -\boldsymbol{J}_{F,j} \boldsymbol{v}_j
\end{equation}
it is possible to formulate each single subsystem optimization. We take the consensus \gls{admm} implementation illustrated in section \ref{subsec::ADMM} and tailor it for the \gls{ocp} in \eqref{eq::ocp}. For convenience, we utilized the scaled dual variable $\Bar{y} = y/\rho$. In this way, the Lagrangian in \eqref{eq::lagrangian} is reduced to its quadratic term. We defined the cost for the subsystem $i$ as:
\begin{equation}
\begin{split}
    \label{eq::lagrangian_dwmpc}
    \Phi^i  =  \boldsymbol{l}_T(\boldsymbol{x_i}(N)) + \sum_{k = 0}^{N-1}{ \Big[ \boldsymbol{l}(\boldsymbol{x}_{k,i} ,\boldsymbol{u}_{k.i})} + \\
    + \sum_{i,j \in\mathcal{X}}||\boldsymbol{r}_{i,j} - \Bar{\boldsymbol{y}}^n_{i,j}||_{\rho}^2 \Big] + ||\boldsymbol{x}_i - \boldsymbol{x}^n||^2_\sigma
    \end{split}
\end{equation}
where the first two terms are the components of the original cost in \eqref{eq::ocp} that depend on the $i^{th}$ subsystem decision variables. $\rho$ and $\sigma$ are positive weight parameters that penalize consensus violations and regularize the solution for the whole-body. The last term of \eqref{eq::lagrangian_dwmpc} introduced by \cite{parallel_admm_convergence} works as a regularizer towards the optimal values found at the previous iteration and it has proven to be crucial for increasing the numerical stability of the algorithm. \\
Thus, the global \gls{ocp} in \eqref{eq::ocp} is reduced to the iterative solution, executed in parallel, of the local \gls{ocp}s:
\begin{equation}  
\label{eq::dwmpc_iterarion}
  \min\limits_{x^i,u^i} \; \Phi^i \quad
 \text{s.t. eq.} \;  \eqref{eq::subsystem_dynamics}\;, \text{eqs.}\;\eqref{sub_eq::ocp_integrator} \; \text{to} \; \eqref{sub_eq::ocp_x0}
%\cref{sub_eq::ocp_integrator,sub_eq::ocp_holo,sub_eq::ocp_z_leg,sub_eq::ocp_uk,sub_eq::ocp_xk,sub_eq::ocp_x0}
\end{equation}
where only the states and control corresponding to the $i^{th}$ subsystem appear as a decision variable. The dynamics constraint in \eqref{sub_eq::ocp_dynamcis} is replaced by \eqref{eq::subsystem_dynamics}, the cost function is the one shown in \eqref{eq::lagrangian_dwmpc}, and the constraint form \eqref{eq::ocp} relative to the subsystem's variables are passed to the local \gls{ocp}. During each iteration of the distributed algorithm, we perform a single full Newton step for each subsystem optimization before updating the dual variable. An outline of the implementation is reported in Algorithm \eqref{algo:dwmpc}, where the \textbf{Map}() function only extracts the subsystem variables from the global one. As for \cite{accelerated_admm} we defined the stopping criteria of our algorithm based on the $l_2$-norm of the residual:
\begin{equation}
    ||r^{n}||_2 < \epsilon
\end{equation}
where $\epsilon$ is a positive scalar threshold on the residual. Our algorithm does not come with convergence guarantees, however, similar to other \gls{admm}-based implementations on nonconvex problems such as those in \cite{accelerated_admm} and \cite{dynamics_consensus}, the solver has empirically proven to be reliable, as shown in Section \ref{subsec::convergence}. 
\begin{figure}
    \centering
    \includegraphics[width=0.48\textwidth]{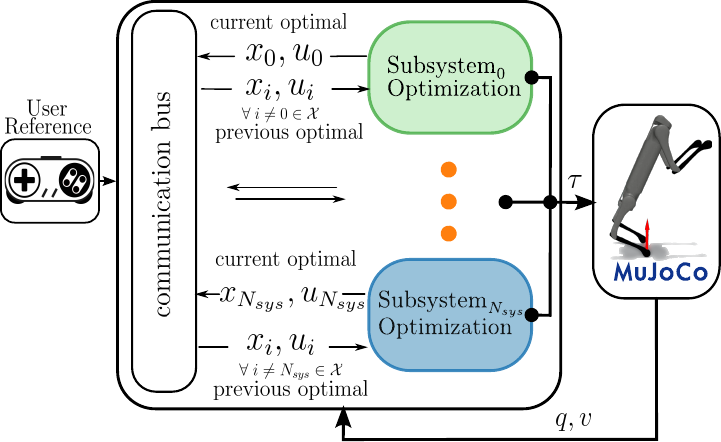}
    \caption{Block scheme of the proposed control framework highlighting the communication bus between the decomposed Subsystem Optimizations.}
    \label{fig:block_scheme}
\end{figure}
\begin{algorithm}[h!]
\caption{Distributed Whole-body Optimization}
\begin{algorithmic}
\State \textbf{Data}: $x^{guess}$, $u^{guess}$
\Repeat
\For{subsystems} in \textit{parallel}
\State $x_i^n$, $u_i^n \longleftarrow$ \textbf{Map}($x^{guess}$, $u^{guess}$)
\State $x_i^{n+1}$, $u_i^{n+1} \longleftarrow$ \textbf{SubSystemSolver}($x_i^n$, $u_i^n$) \eqref{eq::dwmpc_iterarion}
\EndFor
\State $\Bar{y}^{n+1} = \Bar{y}^n + r^{n+1}$
\State $x^{guess}$, $u^{guess} \longleftarrow$ $x^{n+1},u^{n+1}$
\Until stopping criterion
\State
\Return $x^{n+1},u^{n+1}$
\end{algorithmic}
\label{algo:dwmpc}
\end{algorithm}
\subsection{Distributed MPC}
For the online execution of our framework, instead of running the described algorithm to convergence at each control loop, we implemented a receding horizon scheme. In this scheme, each local solver receives the last optimal solution from the other subsystems and the feedback from the robot. Then, we execute one Newton step for each subsystem and apply only the torque calculated at the first step of the horizon to the robot. We then shift the rest of the solution by one time-step and send it back to the communication bus to update the dual variable and provide the current local optimal to the other optimizations for the next control loop. A schematic of the control framework used in the simulation is presented in Fig. \ref{fig:block_scheme}.
\section{Result}
\begin{figure*}[!t]
%\vspace{5pt}
\centering{
\includegraphics[width=0.9\textwidth]{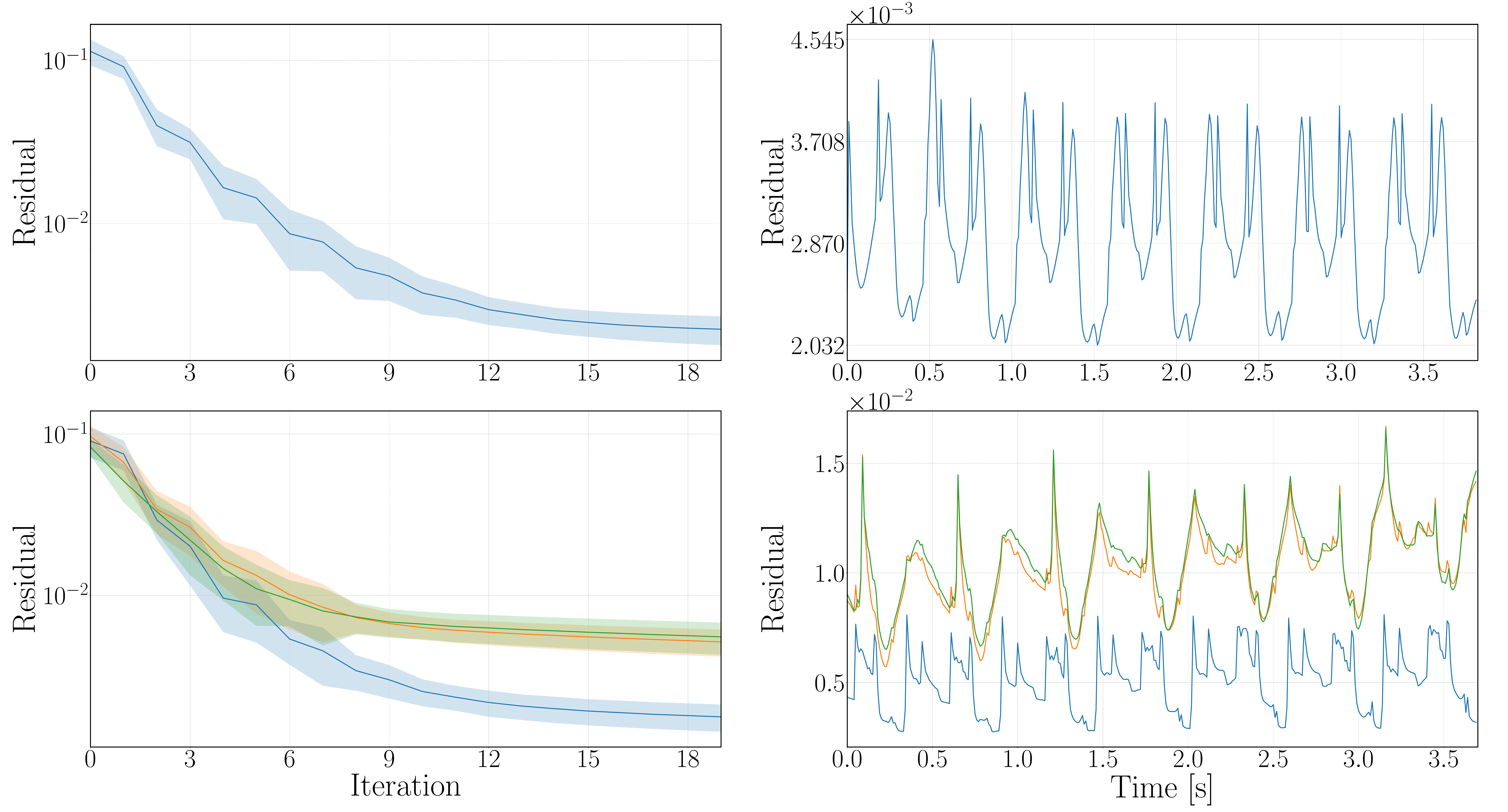}
\vspace{-10pt}%ADL
}    
\caption{The two plots on the left show the trend of the $l_2$ norm of the residuals along the iteration of Algorithm \ref{algo:dwmpc}. On top is the residual for the quadruped with no arm, while on the bottom are the residuals for the quadruped manipulator. The right side plots show the time plot of the residual norm while the robot is trotting in simulation. Again, the top plot is for the robot with no arm, and the bottom one is for the quadruped manipulator.}
\label{fig:residual}
\end{figure*}
\label{sec::result}
In this section, we present the results of different analyses to demonstrate the benefits of the proposed approach. We analyze the performance of our algorithm considering two different systems: a quadruped robot and a quadruped with a manipulator on top, which we will call a \textit{quadruped manipulator}. In the first case, we split the robot into two parts: front and back. All subsystems include a floating base and the decision variables linked with the considered joints, e.g. the front section will consider only the front legs' joint torques, positions, velocities, and the corresponding ground reaction force. This reduces the decision variables per stage from $R^{61}$ to $R^{37}$, since we are discarding the 24 states and control linked with the joints included in the other half of the robot. Thanks to this reduction in the state we can solve the local \gls{ocp} faster, as exemplified by the case of the quadruped manipulator, where the distributed approach can run 4 times faster than the centralized one. Indeed, as depicted in Fig. \ref{fig:solving_time},  for the quadruped manipulator case, the solving time is on average 10 ms for the distributed case, while for the centralized one is 40 ms. 

In the case of the quadruped manipulator, we include a third subsystem in the optimization consisting of the articulated arm, as shown in Fig. \ref{fig:robot_splitted}. In our example, the articulated arm has 6 DoF and thus it adds 18 decision variables per node (joint position, velocity, and torque for each DoF) to the centralized optimization problem. In our formulation, on the other hand, the arm is treated as an independent third optimization with 31 decision variables executed in parallel to the other two. It is worth remembering that the complexity of the optimization scales with the cube of the number of stage decision variables. Indeed, a naive centralized implementation is cursed by the increased dimension of the problem, while our implementation utilizes unexploited computational power by parallelizing the additional subsystem optimization.

In the remainder of this section,  we will first focus on the convergence analysis of the proposed approach, showing the residual trend with both systems and comparing our distributed \gls{mpc} approach with a centralized \gls{mpc} implementation. Subsequently, we will demonstrate the benefits of our approach in terms of solving time and provide examples of agile maneuvers that are possible thanks to the whole-body implementation. The distributed optimization approach and the centralized one have been implemented using Adam \cite{adam}, an open-source library for rigid-body dynamics and Acados \cite{acados}, an open-source Sequential Quadratic Programming solver tailored for optimal control. We chose to utilize Acados for its ease of use, but our method is not tailored to any specific solver and can be reused with other \gls{nlp} solvers. Both formulations utilize a prediction horizon of 50 nodes with a discretization time of 0.01 seconds, and they are both run at 50Hz for the sake of comparison. All the presented results have been obtained using an Intel Core i7 13700H laptop CPU. For the simulations, we utilized Mujoco \cite{mujoco} as physics engine.

\subsection{Convergence analysis}

\begin{figure}[!t]
\centering{
\includegraphics[width=0.4\textwidth]{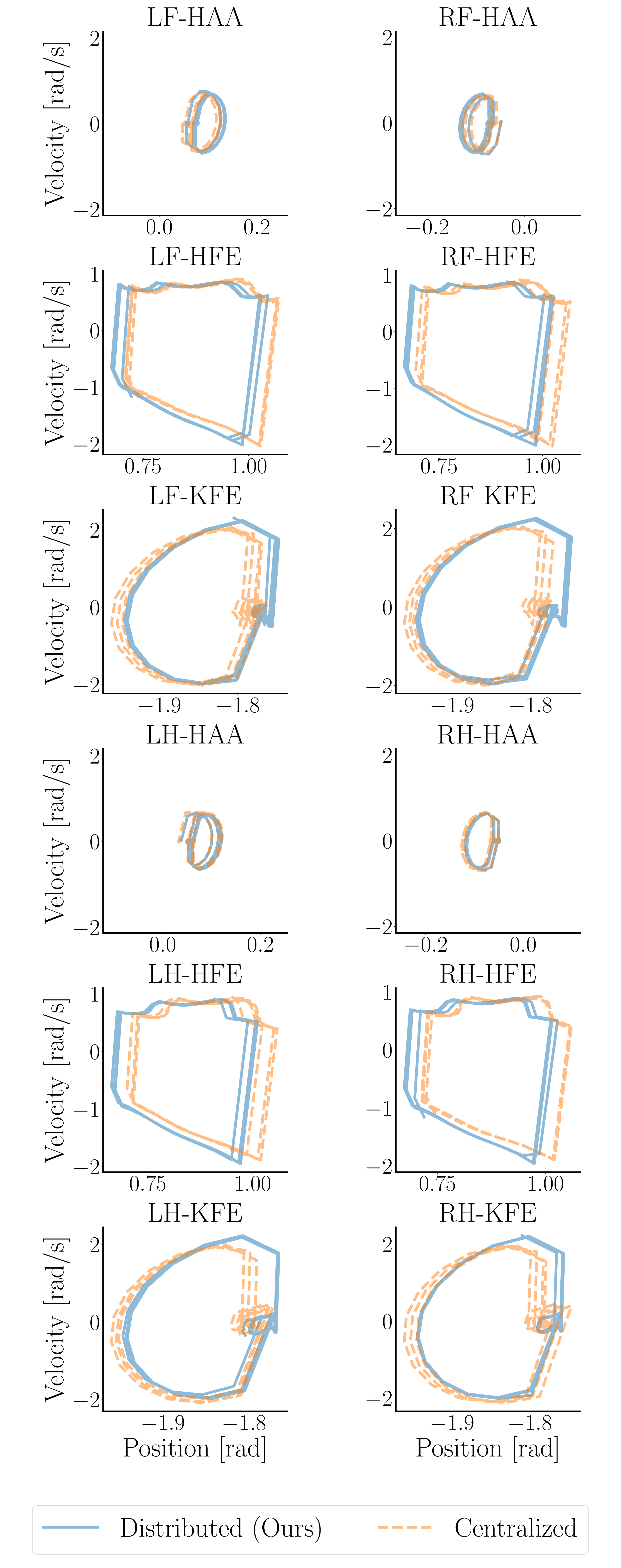}
}    
\caption{Phase plots of the robot trotting with the centralized (dashed blue line) and distributed (orange line) solutions. The plot shows a recording of the robot trotting in simulation at a desired speed of 0.3 $m/s$. Where RF-LF-LH-RH stands for Right Front, Left Front, Left Hind, Right Hind and, HAA, HFE, KFE stand for Hip Abduction/Adduction, Hip Flexion/Extension, and Knee Flexion/Extension.}
\vspace{-18pt}%ADL
\label{fig:comarison}
\end{figure}
\label{subsec::convergence}
In this subsection, we aim to empirically analyze the convergence properties and stability of our algorithm. We begin by examining the $l_2$ norm of the residual, as calculated in \eqref{residual}. This metric provides insight into the quality of the consensus achieved. Fig. \ref{fig:residual} illustrates the trend over the iterations of Algorithm \ref{algo:dwmpc}. As previously explained, we consider two scenarios: one with the quadruped split in half and the other with the robot plus the arm. It should be noted that in the former case, the residual is one, while in the latter case, they become 3, indicating consensus among the three subsystems. The plots depict the average over one hundred simulations, with the system initialized in different configurations and with randomized warm starts. The variance is represented by the colored area around the mean value. As shown, the residual converges within a few iterations to small values. Notably, the value the residual reaches corresponds to an error between the subsystems at the end of the predicted trajectories in the orders of millimeters or less. In the second column of Fig. \ref{fig:residual}, we display the trend of the residual norm of the \gls{mpc}. In this case, we reported how the residual value remains in the order of $10^{-3}$ when the distributed approach is utilized in a receding horizon fashion. The data was recorded while the robot was trotting in simulation following a desired speed of $0.3\frac{m}{s}$. 

Finally, in Fig. \ref{fig:comarison} we compare, in simulation, the centralized and the distributed approach, reporting the phase plot for each joint. The figure illustrates that the steady-state solutions of both methods converge close to the same values, depicting how our solver response is the same as one from a centralized implementation.

\subsection{Performance evaluation}
\begin{figure}[!t]
\centering{
\includegraphics[width=0.45\textwidth]{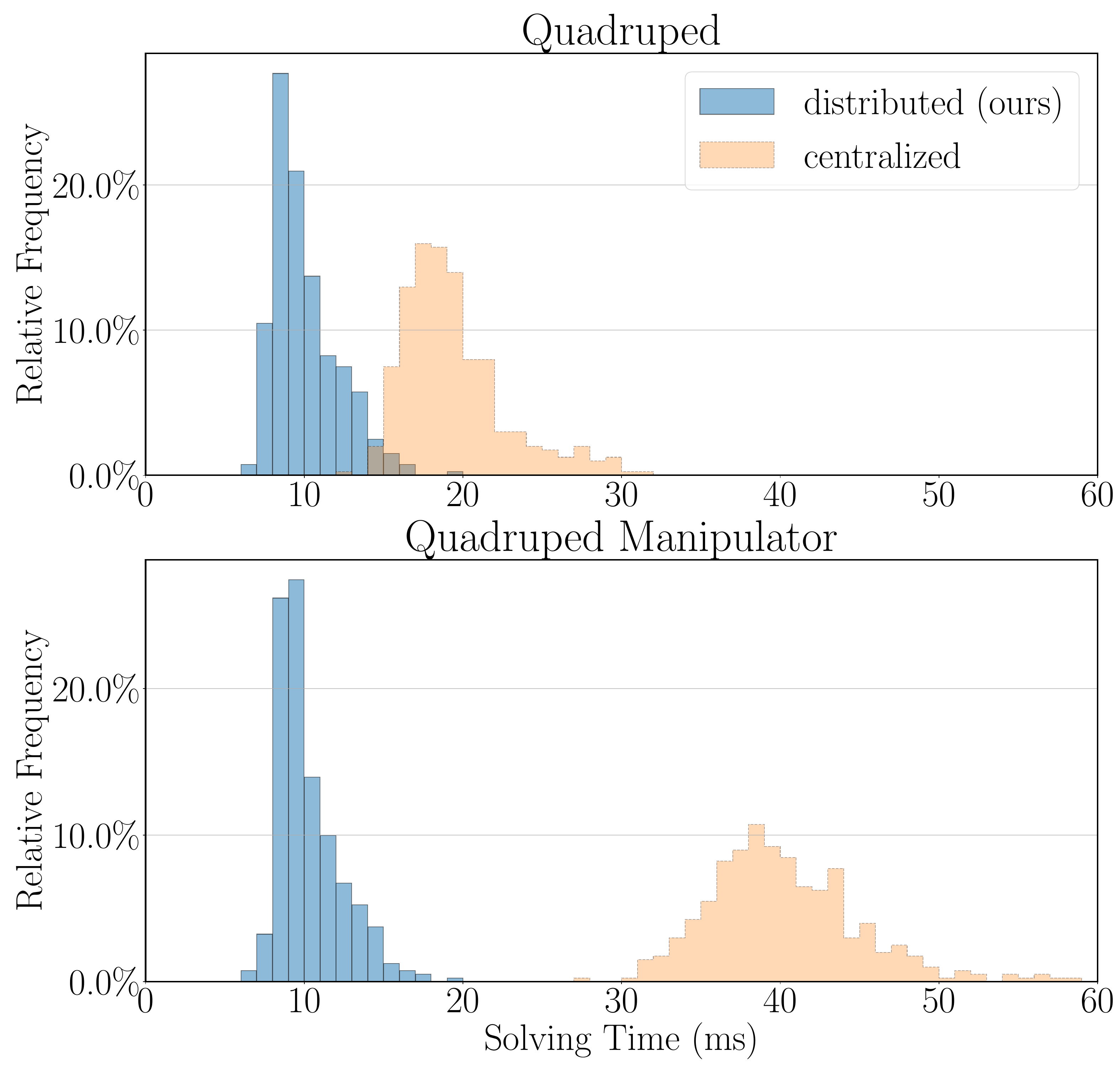}
}    
\caption{Relative frequencies for the occurrence of the solution time of the receding horizon problem in \eqref{eq::ocp} recorded over one hundred simulations. In blue is our distributed implementation while in orange is the centralized one. On the top, the solution refers to the quadruped robot model while on the bottom plot the quadruped plus the arm.}
\label{fig:solving_time}
\vspace{-18pt}%ADL
\end{figure}
\label{subsec::performance}
In this subsection, we show the main benefits of our implementation. As displayed in Fig. \ref{fig:solving_time} we compared the solving time of the distributed and centralized implementation for the quadruped and quadruped manipulator. We showed the occurrence of the solving times recorded in a 5 second simulation reporting in the histogram the relative frequencies. In simulation, the robot is asked to follow a desired velocity of $0.3 m/s$ and is perturbed repeatedly by an external force of random direction and magnitude in the range between $\pm30N$ for 0.1s each time. The graph shows that the solving times for the distributed controller are on average half of the ones of the centralized implementation for the quadruped robot while for the quadruped manipulator, the solving time is reduced by a factor of four. 

It is worth noting that our implementation is not affected by the increased complexity of the system. Adding the arm on the quadruped robot does not increase the computational time. It should be mentioned that the times reported for the distributed implementation are the one of the slowest subsystem optimizations. This consideration remains valid, as long as the new system we want to include can be considered as an additional subsystem that can be processed in parallel as it has been done for the arm. As already mentioned, even though the reported times refer to our implementation using Acados,  changing the solver at the core of our implementation will not affect the relative gains, but only the absolute numbers in the same way for both centralized and distributed approaches. 

Finally, to demonstrate the benefit of a whole-body implementation and the robustness of our formulation, we show the robot performing complex maneuvers that need whole-body coordination. In the top part of Fig. \ref{fig:agility} we show the quadruped stand-up and walk on two feet. No particular effort is required in designing the reference trajectories since the \gls{mpc} can generate and stabilize agile maneuvers starting from simple unfeasible trajectories. Indeed, for the emergence of the biped walking behavior, it was sufficient a step function for the robot's desired base pitch equal to $\pi/2$ and a reference constant velocity to the base equal to $0.2 m/s$.

The bottom part of Fig. \ref{fig:agility} shows the quadruped manipulator trotting while its end-effector is tracking a triangular spiral defined in the world frame. In this case, the only reference we are passing to the \gls{mpc} is the desired spiral trajectory for the arm's end-effector.

Both simulations can be seen in the accompanying video.

%\begin{figure*}[!t]
%\vspace{5pt}
%\centering{
%\includegraphics[width=0.98\textwidth]{figure/stand_up.pdf}
%}    
%\caption{Snapshot of the quadruped standing up on two feet and walking forward}
%\label{fig:stand_up}
%\end{figure*}
%\begin{figure}[!t]
%\vspace{5pt}
%\centering{
%\includegraphics[width=0.48\textwidth]{figure/arm.pdf}
%}    
%\caption{Snapshot of the quadruped manipulator following a triangular spiral}
%\label{fig::arm}
%\end{figure}

\section{Conclusion}
\label{sec::conclusion}
This paper presented a novel approach to decompose the robot dynamics into reduced and more tractable subsystems to accelerate \gls{mpc} for legged robots. Utilizing \gls{admm} we ensure consensus among the parallel subsystem's optimization demonstrating significant improvements in computational efficiency. The parallelizability of our method facilitates the integration of additional limbs such as articulated arms without compromising the solving time, and thus making the system easily scalable. Through extensive numerical simulations, we have validated the convergence and performance of our approach across two systems of varying complexity, highlighting its effectiveness compared to state-of-the-art whole-body \gls{mpc} implementations.
The quantitative simulations have underscored the substantial reduction in computational time; in particular, for the quadruped model we gained a reduction of two times while for the quadruped manipulator, the achieved reduction is four times. These gains derive from the fact that we have added a new subsystem to be solved in parallel with the others so the computational time for our approach remains constant, while for a centralized whole-body optimization it drastically increases due to a higher number of decision variables per node.
These gains in computational time come from our more effective use of modern hardware resources.

As future work, we plan to test our implementation on hardware and include other popular nonlinear solvers such as \cite{crocoddyl} and \cite{ocs2} in our framework allowing to choose the most suitable solver for a given scenario. We are also going to extend our work to other robot morphologies, e.g. humanoid robots.
\bibliographystyle{IEEEtran}
\bibliography{references/references}

% that's all folks
\end{document}